\pdfoutput=1

\documentclass[11pt]{article}

\usepackage[preprint]{acl}

\usepackage{times}
\usepackage{latexsym}

\usepackage[T1]{fontenc}

\usepackage[utf8]{inputenc}

\usepackage{microtype}

\usepackage{inconsolata}

\usepackage{tabularx, booktabs}
\usepackage{array}
\usepackage{url}
\usepackage{graphicx}
\usepackage{enumitem}

\newcommand{\dataset}{\textsc{Fieldwork}}
\newcommand{\task}{\textsc{wav2gloss}}

\title{\task{}: Generating Interlinear Glossed Text from Speech}

\author{
\textbf{Taiqi He\textsuperscript{1}},
\textbf{Kwanghee Choi\textsuperscript{1}},
\textbf{Lindia Tjuatja\textsuperscript{1}},
\textbf{Nathaniel R. Robinson\textsuperscript{2}},
\textbf{Jiatong Shi\textsuperscript{1}},
\\
\textbf{Shinji Watanabe\textsuperscript{1}},
\textbf{Graham Neubig\textsuperscript{1}},
\textbf{David R. Mortensen\textsuperscript{1}},
\textbf{Lori Levin\textsuperscript{1}}
\\
 \textsuperscript{1}Language Technologies Institute, Carnegie Mellon University \\
 \textsuperscript{2}Center for Language and Speech Processing, Johns Hopkins University
}

\begin{document}
\maketitle
\begin{abstract}

Thousands of the world's languages are in danger of extinction---a tremendous threat to cultural identities and human language diversity. Interlinear Glossed Text (IGT) is a form of linguistic annotation that can support documentation and resource creation for these languages' communities. IGT typically consists of (1) transcriptions, (2) morphological segmentation, (3) glosses, and (4) free translations to a majority language. We propose \task{}: a task in which these four annotation components are extracted automatically from speech, and introduce the first dataset to this end, \dataset{}:\footnote{\url{https://huggingface.co/datasets/wav2gloss/fieldwork}} a corpus of speech with all these annotations, derived from the work of field linguists, covering 37 languages, with standard formatting, and train/dev/test splits. We provide various baselines to lay the groundwork for future research on IGT generation from speech, such as end-to-end versus cascaded, monolingual versus multilingual, and single-task versus multi-task approaches.

\end{abstract}

\section{Introduction}
\label{sec: intro}

Working against the overwhelming tide of social and historical forces, linguists and community activists from around the world have set out to record endangered languages while they are still actively spoken. As a first step, these documentary efforts involve---quite literally---recordings, then transcriptions, translations, and other annotations. The ultimate goal of such efforts is often to take a large volume of recorded speech and annotate it with Interlinear Glossed Text (IGT).

\begin{figure}[t]
  \centering
  \includegraphics[width=0.9\linewidth]{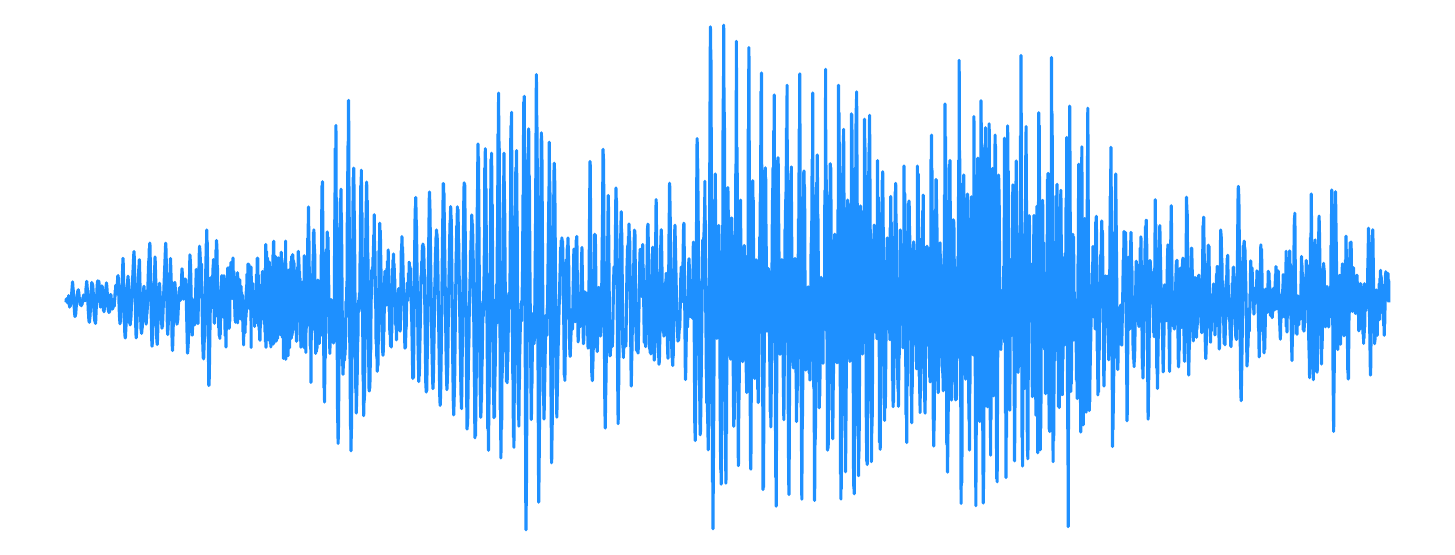}
  \vspace{1ex}
\begin{tabular}{rlllll}
  
  \texttt{wd}: & n & \multicolumn{2}{l}{s\`iginde} & yan & de \\
  \texttt{sr}: & n & s\`igi & -nde                 & yan & de \\
  \texttt{ur}: & n & s\`igi & -len                 & yan & le \\
  \texttt{gl}: & 1.SG & sit & -PC.RES            & that & FOC \\
  \texttt{tr}: & \multicolumn{5}{l}{``I live here.''} \\
\end{tabular}
  \caption{A representation of a single Kakabe utterance in the \dataset{} corpus: speech paired with annotations. \texttt{wd} is the unsegmented transcription; \texttt{ur} and \texttt{sr} are the underlying and surface representations; \texttt{gl} is a morpheme-by-morpheme gloss; and \texttt{tr} is a free translation into the metalanguage.}
  \label{fig:kakabe-example}
\end{figure}

IGT is the \textit{lingua franca} of documentary linguistics. Most IGT now follows a set of conventions called the Leipzig glossing rules \citep{comrie2008leipzig} but other formats are in use. An example of IGT from \citet{doreco-kaka1265} is shown in Figure~\ref{fig:kakabe-example}. It consists of an unsegmented transcription (\texttt{wd}), underlying (\texttt{ur}) and surface forms (\texttt{sr}) segmented into morphemes, morpheme labels (glosses; \texttt{gl}), and free translation (\texttt{tr}) aligned with one another and the source audio recording. The lines labeled \texttt{sr}, \texttt{ur}, and \texttt{gl} are most important to linguists and language teachers. Together, they tell the user how linguistic form maps into linguistic function, enabling the creation of many valuable resources and a variety of useful analyses.

Most linguistic field recordings, though, never make it to IGT \citep{SeifartEvansHammarstromEtAl2018}. Simply transcribing field data (without other annotations) can take up to one hour per minute of recorded speech \citep{do:halshs-00980431}. Adding additional annotations is even more expensive. This bottleneck keeps vast collections of field recordings from achieving their full documentary potential. 

Producing IGT from audio is a tractable problem. While linguists can do little or nothing to address the underlying causes of language endangerment---which form an intersecting lattice of political, cultural, and economic factors---speech and natural language processing researchers can do even less. Technologists can, however, facilitate the efforts of field linguists and language workers to \textit{document} endangered languages by developing technologies that make the mammoth tasks of annotating field data surmountable \citep{shi-etal-2021-leveraging, shi-etal-2021-highland}. For example, they can develop models that automatically generate first-pass transcriptions from raw speech. Research has demonstrated that such models can speed up transcription dramatically \cite{amith2021end}.

In direct support of the goal of language documentation, we propose a new speech and language processing task: \task{}. This task assumes recorded speech as the only input. The output consists of aligned annotations for transcription (with and without segmentation), glossing, and translation. In order to allow the research community to participate in this task, we introduce the following:
\begin{enumerate}[nosep]
\item The \dataset{} Corpus, a speech+IGT dataset for 37 languages---drawn from five archives of linguistic field data---with a standard format and train/dev/test splits.
\item Four subtasks crucial for language documentation: prediction of transcription, underlying representation, gloss, and translation. To our knowledge, this is the first attempt to extract these annotations directly from speech. 
\item Benchmarks based on well known speech and NLP models, including end-to-end and cascaded approaches to predicting IGT from speech.
\end{enumerate}

\begin{figure*}[t]
  \centering
  \includegraphics[width=0.9\textwidth]{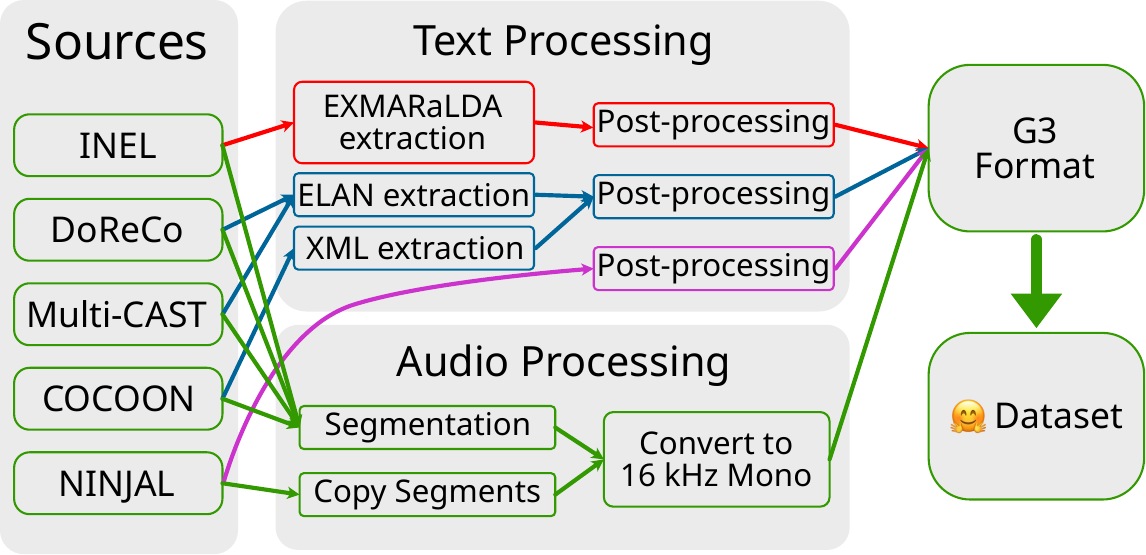}
  \caption{
  A visualization of the building of the \dataset{} dataset.
  }
  \label{fig:data-processing}
\end{figure*}

\section{Dataset}

\begin{table*}[]
    \centering
    \begin{tabularx}{\textwidth}{lllccc}
    \toprule
    \textbf{Glottocode} & \textbf{Name} & \textbf{CC Type} & \textbf{Train (h)} & \textbf{Dev+Test (h)} \\
    
    \midrule
    \multicolumn{3}{l}{\textbf{DoReCo}} \\
    \cmidrule(r){1-1}
    \texttt{beja1238} & Beja \cite{doreco-beja1238} & BY-NC & 1.55 & 0.29 \\
    \texttt{ruul1235} & Ruuli \cite{doreco-ruul1235} & BY & 0.96 & 0.28 \\
    \texttt{texi1237} & Texistepec Popoluca \cite{doreco-texi1237} & BY & 0.84 & 0.26 \\
    \texttt{komn1238} & Komnzo \cite{doreco-komn1238} & BY & 0.73 & 0.42 \\
    \texttt{arap1274} & Arapaho \cite{doreco-arap1274} & BY & 0.56 & 0.88 \\
    \texttt{goro1270} & Gorwaa \cite{doreco-goro1270} & BY & 0.52 & 0.45 \\
    \texttt{teop1238} & Teop \cite{doreco-teop1238} & BY & 0.52 & 0.52 \\
    \texttt{nngg1234} & N||ng \cite{doreco-nngg1234} & BY & 0.52 & 0.33 \\
    \texttt{sumi1235} & Sümi \cite{doreco-sumi1235} & BY & 0.40 & 0.40 \\
    \texttt{jeju1234} & Jejuan \cite{doreco-jeju1234} & BY & 0.38 & 0.65 \\
    \texttt{bora1263} & Bora \cite{doreco-bora1263} & BY & 0.23 & 1.44 \\
    \texttt{apah1238} & Yali (Apahapsili) \cite{doreco-apah1238} & BY-NC-SA & 0.18 & 0.27 \\
    \texttt{port1286} & Daakie \cite{doreco-port1286} & BY & 0.14 & 0.75 \\
    \texttt{savo1255} & Savosavo \cite{doreco-savo1255} & BY & 0.10 & 1.20 \\
    \texttt{trin1278} & Mojeño Trinitario \cite{doreco-trin1278} & BY & - & 1.56 \\
    \texttt{sout2856} & Nafsan (South Efate) \cite{doreco-sout2856} & BY-NC-SA & - & 1.55 \\
    \texttt{pnar1238} & Pnar \cite{doreco-pnar1238} & BY-NC & - & 0.91 \\
    \texttt{kaka1265} & Kakabe \cite{doreco-kaka1265} & BY & - & 0.90 \\
    
    \midrule
    \multicolumn{3}{l}{\textbf{Multi-CAST}} \\
    \cmidrule(r){1-1}
    \texttt{vera1241} & Vera'a \cite{vera1241} & BY & 1.02 & 0.97 \\
    \texttt{tond1251} & Tondano \cite{tond1251} & BY & 0.22 & 0.67 \\
    \texttt{taul1251} & Tulil \cite{taul1251} & BY & - & 1.18 \\
    \texttt{arta1239} & Arta \cite{arta1239} & BY & - & 0.91 \\
    \texttt{nort2641} & Northern Kurdish \cite{nort2641} & BY & - & 0.86 \\
    \texttt{tehr1242} & Persian \cite{tehr1242} & BY & - & 0.82 \\
    \texttt{taba1259} & Tabasaran \cite{taba1259} & BY & - & 0.79 \\
    \texttt{sanz1248} & Sanzhi Dargwa \cite{sanz1248} & BY & - & 0.67 \\
    \texttt{kach1280} & Jinghpaw \cite{kach1280} & BY & - & 0.66 \\
    \texttt{mand1415} & Mandarin \cite{mand1415} & BY & - & 0.66 \\
    \texttt{sumb1241} & Sumbawa \cite{sumb1241} & BY & - & 0.63 \\
    \texttt{kara1499} & Kalamang \cite{kara1499} & BY & - & 0.59 \\

    \midrule
    \multicolumn{3}{l}{\textbf{COCOON}} \\
    \cmidrule(r){1-1}
    \texttt{slav1254} & Slavomolisano \cite{BreuPiccoliEtAl18} & BY-NC & 1.01 & 0.96 \\
    \texttt{balk1252} & Balkan Romani \cite{adamou2015} & BY-NC-SA & - & 0.35 \\
   
    \midrule
    \multicolumn{3}{l}{\textbf{INEL}} \\
    \cmidrule(r){1-1}
    \texttt{dolg1241} & Dolgan \cite{inel-dolgan} & BY-NC-SA & 11.64 & 1.23 \\
    \texttt{kama1378} & Kamas \cite{inel-kamas} & BY-NC-SA & 9.91 & 1.15 \\
    \texttt{selk1253} & Selkup \cite{inel-selkup} & BY-NC-SA & 1.70 & 1.15 \\
    \texttt{even1259} & Evenki \cite{inel-evenki} & BY-NC-SA & 1.54 & 1.13  \\
    
    \midrule
    \multicolumn{3}{l}{\textbf{NINJAL}} \\
    \cmidrule(r){1-1}
    \texttt{ainu1240} & Ainu \cite{ninjal-ainu-folklore} & BY-SA & 7.12 & 1.13 \\
    \midrule
     & \dataset{} Total & BY-NC-SA & 41.79 & 29.56 \\
    \bottomrule
    \end{tabularx}
    \caption{Overview of languages included in the \dataset{} dataset. All licenses are CC with the specific restrictions for each language listed. We also show the hours of training data and combined dev and test data available for each language.}
    \label{tab:languages}
\end{table*}

We present the \dataset{} dataset, a collection of linguistic field recordings with audio that has been transcribed and glossed in IGT. We build upon DoReCo \citep{doreco:2022} and Multi-CAST, which are curated collections of field data, as well as data released through the COCOON repository,\footnote{\url{https://cocoon.huma-num.fr/exist/crdo?lang=en}} and data produced by the INEL project\footnote{\url{https://www.slm.uni-hamburg.de/en/inel.html}} and NINJAL \citep{ninjal-ainu-folklore}. Our main contributions are selecting data for which there is both audio and gloss; compiling this data into a single structured, computer accessible dataset; and providing transcription and glossing benchmarks for each language in the dataset. Our work would not have been possible without the dedicated work of expert field linguists and speaker communities.

Our first step in adapting the \dataset{} corpora for our four subtasks is to select languages where both audio and IGT are available. There are some overlaps between DoReCo, Multi-CAST, and INEL. For each language that appears more than once in those three sources, we only use the data from the source that contains the highest number of utterances.
We do not attempt to merge corpora of the same language from multiple sources. We also require all languages we select to have a permissive CC license without a No Derivatives (ND) restriction.\footnote{We base our decision on a layperson's reading of the Creative Commons licenses, which counts cleaned, reformatted, and standardized versions of datasets as derivative works.} The list of language corpora selected along with their license information, and the number of hours of data available in training and combined dev and test splits is shown in Table \ref{tab:languages}. See Figure \ref{fig:data-processing} for an overview of the data processing pipelines.

Annotated data come in a variety of formats such as JSON or XML. Most of our source data come in XML-based formats---with ELAN \citep{elan} being the most popular---as well as EXMARaLDA \citep{exmaralda} and Pangloss DTD \citep{pangloss-dtd}.\footnote{See \citet{von-prince-nordhoff-2020-empirical} for a more detailed overview of ELAN and common annotator practices.}
In ELAN and EXMARaLDA, annotations are text strings with beginning and end time stamps, organized into tiers including underlying form, surface form, transcription, gloss, and unique ID.

We extract annotations associated with utterances along with their corresponding audio spans, convert all audio files to WAV format with a single channel and 16 kHz sampling rate, and store annotations in an intermediate YAML based format \citep{mortensen-etal-2023-generalized} that is easier to process, read, and edit. We manually inspect the annotations for conversion errors and non-speech markers.

Finally, we partition the corpus into train/dev/test splits. We create partitions that contain full documents in order to preserve the contextual information. 
Assuming each document covers a different topic and has slightly different recording conditions, using full documents will make modeling more challenging and realistic, since the dev and test splits will be out-of-distribution, with minimal overlap in content between the splits. 
For each language, we look at the number of utterances in total to determine the splits. 
If there are fewer than 200 utterances, all of them are assigned to the test set; if there are between 200 and 1,000 utterances, we assign 25\% of the data to the dev set, and the rest to the test set; if there are more than 1,000 utterances, we assign 250 utterances to the dev set, 750 utterances to the test set, and the rest to the training set. 
With those partitions determined, we use a knapsack solver\footnote{\url{https://github.com/google/or-tools}} to optimally assign documents to splits based on the number of utterances within each document.\footnote{Because many of our datasets contained only 1-2 speakers, it was impossible to avoid speaker overlap between splits. However, this is not as serious concern for our application as it may be for others, since language documentation is almost always conducted with a small number of speakers.} 
We then apply final text cleaning by normalizing punctuations and removing special symbols on the transcriptions and translations, and convert the corpora into a Hugging Face dataset which can be readily used to train and test speech-to-text models.

In the following paragraphs, we present information on the specific data sources and unique processing required for each.

\paragraph{DoReCo}
\textbf{Do}cumentation \textbf{Re}ference \textbf{Co}rpus \citep{doreco-desc} provides time-aligned transcriptions for 51 under-resourced languages. 
\citeauthor{doreco-desc} processed each language corpus through a pipeline with consistency checking, multiple rounds of audio-text alignment, and manual corrections. We selected only data that is marked as fully (vs. partially or not) glossed.  The annotations in the DoReCo dataset are in the ELAN format and organized such that transcriptions, translations, and utterance IDs share the same time span at the utterance level, and the underlying forms and glosses share the time span at the morpheme level within utterance time spans. After extracting the annotations, we drop words or utterances that consist only of the pause marker \texttt{<p:>}, and replace each span marked as \texttt{<label<text>\/>} with its \texttt{text} component.

\paragraph{Multi-CAST}
\textbf{Multi}lingual \textbf{C}orpus of \textbf{A}nnotated \textbf{S}poken \textbf{T}exts is another collection of annotated time-aligned speech \citep{multicast}. Similar to DoReCo, Multi-CAST contains annotated speech for 18 languages, with robust annotation guidelines \citep{haig2015annotations}. The annotations are also stored in the ELAN format; thus, data extraction is similar to that for DoReCo. One notable difference is Multi-CAST's designation of non-speech annotations, which start with \texttt{\#}, \texttt{0}, or \texttt{\%}. We delete text tokens marked with these non-speech symbols from the underlying form tier and the gloss tier.

\paragraph{INEL}
Grammars, Corpora and Language Technology for \textbf{I}ndigenous \textbf{N}orthern \textbf{E}urasian \textbf{L}anguages is an ongoing project at the Academy of Sciences and Humanities in Hamburg and the University of Hamburg that focuses on gathering resources for indigenous languages and language varieties of Northern Eurasia.\footnote{\url{https://www.slm.uni-hamburg.de/en/inel.html}} We use all four of the languages released so far (see Table \ref{tab:languages}) and use all annotated speech available. The datasets are annotated with EXMARaLDA \citep{schmit2014exmaralda}, which is structurally similar to ELAN.

\paragraph{COCOON}
\textbf{Co}llections de \textbf{Co}rpus \textbf{O}raux \textbf{N}umériques is a large repository of field linguistic data that contains a variety of data types from a wide range of researchers.\footnote{\url{https://cocoon.huma-num.fr}} To narrow down the data that are of interest to us, we first obtained a list of annotation files within the archive through the OLAC Aggregator,\footnote{Accessible through \url{http://www.language-archives.org/cgi-bin/olaca3.pl?verb=Document}. Our data is up-to-date as of Feb 23, 2023. In case the aggregator is not available, metadata can also be obtained through the CLARIN OAI harvester: \url{https://github.com/clarin-eric/oai-harvest-manager}.} targeting the Pangloss DTD format used specifically within COCOON. After the list of annotation files was obtained, we retrieved them along with associated media files, and sorted the results by language. We then used a simple heuristic---checking that there were multiple levels of word-level annotation and that morpheme-level annotation was available---to select a subset of data that likely contained speech with IGT. We then did a first round of manual verification to make sure the license information was available and suitable, and then a second round of more detailed checks for IGT quality. Two languages remained after the process, as shown in Table~\ref{tab:languages}. Though a third language, Kakabe, would also fit our criteria, we chose to use its DoReCo corpus because COCOON data are generally noisier and are not automatically aligned and manually checked as DoReCo data are.

\paragraph{NINJAL Ainu Folklore}
Ainu is a nearly extinct language spoken in Hokkaido, Japan. The \textbf{NINJAL} Glossed Audio Corpus of \textbf{Ainu Folklore} \citep{ninjal-ainu-folklore} contains recordings of 38 traditional Ainu folktales by two Ainu speakers, along with their transcriptions (in Latin script and occasionally Japanese script), English and Japanese translations, and underlying and surface gloss forms in English and Japanese. We used the Latin transcriptions and English translation/glosses. We scraped data via the corpus's web interface\footnote{\url{https://ainu.ninjal.ac.jp/folklore/en/}} and stored them in their native JSON format. We communicated with the authors and obtained permission to share them.

\section{Experiments}

We provide benchmarks for automatically generating IGT by fine-tuning pre-trained speech and text models.
We choose commonly used models, methods, and finetuning settings with their corresponding codebases to provide a solid baseline for future research. See Appendix \ref{apdx:params} for details on models sizes and hyper-parameter settings.

\subsection{End-to-End models}\label{subsec:e2e}

Three of our four tasks (transcription, underlying form, and IGT prediction) are monotonic sequence-to-sequence tasks, similar to automatic speech recognition (ASR).
Hence, we employ standard ASR training methods for prediction of each annotation (transcription, underlying form, gloss, and translation).
Even though translation is not monotonic with respect to time as the other tasks are, we use the same training scheme, since previous work has found that multi-head attention-based networks can implicitly model non-monotonicity \cite{yan2023ctc}.
Meanwhile, by using the same scheme, we can provide a more straightforward comparison between the performance of different tasks and approaches.

For end-to-end approaches, we use ESPnet \cite{watanabe2018espnet} to employ two representative families of state-of-the-art pre-trained speech models: self-supervised and semi-supervised.
The first self-supervised model we employ is WavLM Large \cite{chen2022wavlm}, which achieves state-of-the-art performance on the SUPERB benchmark, a leaderboard for various speech-related tasks \cite{yang21c_interspeech, feng2023superb}.
We also use XLS-R-300M \cite{babu2021xls}, a model specifically trained for cross-lingual capabilities, which has shown superior performance on the ML-SUPERB benchmark in multilingual tasks \cite{shi23g_interspeech, shi2023findings}.
WavLM and XLS-R are also in the family of HuBERT- and wav2vec2.0-like models \cite{hsu2021hubert,baevski2020wav2vec}, respectively, which are commonly used self-supervised models. 
Similar to \citet{chen2023improving}, while we freeze the self-supervised models to preserve acoustic knowledge, we attach a conformer encoder and transformer decoder \cite{guo2021recent} to support the four different annotations we infer.
Both the encoder (50M parameters) and decoder (26M parameters) have six blocks and eight attention heads, with the addition of the 315M frozen parameters from each of the pre-trained models, bringing the total parameter count to 391M. WavLM was pretrained on English only, while XLS-R was pretrained on multiple languages, two of which---Mandarin and Persian---are also in \dataset{}.
We train the model with CTC-Attention loss \cite{kim2017joint}, where CTC and attention loss are applied to the encoder and decoder, respectively.
We train the language model from scratch and employ character-level tokenization with added  language and task tokens.
For details, see our public source code.\footnote{\url{https://github.com/juice500ml/espnet/tree/wav2gloss/egs2/wav2gloss/asr1}}

On the other hand, supervised speech models, such as Whisper \cite{radford2023robust} or OWSM \cite{peng2023reproducing}, show reasonably robust performance in various tasks, especially ASR.
Hence, we fine-tune the OWSM-v3.1-base model (\citealt{peng2024owsm}, 101M parameters), which is an open-sourced reproduction of Whisper with public training software and datasets. 
We use OWSM in our experiments because its open-source nature is desirable for reproducibility and thorough scientific analysis. 
These supervised models already contain a predefined vocabulary and the corresponding tokenizer.
OWSM uses a BPE tokenizer with 50k vocabulary, trained on a random subset of its training data. 
We add a specific token per language and two task tokens for gloss and underlying forms.
For transcription and translation, we utilize the existing task tokens. The pretraining corpus of OWSM is also multilingual, with Mandarin and Persian being the only two overlapping languages with \dataset{}.
Just as with self-supervised models, we fine-tune with the CTC-Attention loss (the same approach as in OWSM pre-training). 
Similar to \citet{rouditchenko23_interspeech}, we fully fine-tune the OWSM model.
The source code for fine-tuning OWSM is available in a public repository.\footnote{\url{https://github.com/juice500ml/finetune_owsm}}

We train a small number of OWSM models monolingually for transcription in each language, for comparison in Section \ref{sec:discussion}.
The remainder we train in a multilingual manner, including all the languages in \dataset{}.
During training, we evaluate the models using the sample-wise average accuracy on the dev set after each epoch, and keep the checkpoint with the highest accuracy.
We experiment with both single-task and multi-task models: in the single-task paradigm, we train an individual model for each of the output forms (transcription, underlying, gloss, translation); in the multi-task paradigm, we train a single multi-task model that predicts different output forms based on the task token.
We compare the performance in Section \ref{sec:results}.

\subsection{Cascaded model}
From our preliminary experiments, we see that predicting glosses from speech is much more challenging than predicting transcriptions (i.e. ASR).
The 2023 SIGMORPHON shared task on interlinear glossing \citep{ginn-etal-2023-findings} showed that text-to-gloss models can achieve over 90\% gloss prediction accuracy in certain languages, given enough training data. 
Therefore, we evaluate a cascaded approach where we take the best performing ASR models in the end-to-end setting, and use their transcription outputs as inputs to a text-to-gloss model.
We use two text models initialized from ByT5-base (\citealt{xue2022byt5}, 582M parameters), one trained only on \dataset{} for the underlying form, gloss, and translation tasks; the other fine-tuned first on ODIN \citep{lewis-xia-2010} and then fine-tuned on \dataset{} for gloss and translation, similar to the approach of \citet{he-etal-2023-sigmorefun}. All text models are single-task, meaning we train a separate model for each task within each setting. 
While the ByT5 model was not the best performing in the shared task, it can easily support multilingual training and does not require the inputs to be segmented, and therefore is more suitable for the unsegmented outputs of the ASR systems.

\begin{table*}[t]
    \centering
    \begin{tabular}{l *{8}{c}}
        \toprule
        & \multicolumn{2}{c}{\textbf{Transcription}} & \multicolumn{2}{c}{\textbf{Underlying}} & \multicolumn{2}{c}{\textbf{Gloss}} & \multicolumn{2}{c}{\textbf{Translation}} \\
        & \multicolumn{2}{c}{CER $\downarrow$} & \multicolumn{2}{c}{CER $\downarrow$} & \multicolumn{2}{c}{CER $\downarrow$} & \multicolumn{2}{c}{chrF++ $\uparrow$} \\
        \cmidrule(lr){2-3} \cmidrule(lr){4-5} \cmidrule(lr){6-7} \cmidrule(lr){8-9}
        \textbf{Model} & Seen & Unseen & Seen & Unseen & Seen & Unseen & Seen & Unseen \\
        \midrule
        \multicolumn{8}{l}{\textbf{Multi-task}} \\
        \cmidrule(r){1-1}
        WavLM E2E & 76.9 & 77.8 & 66.3 & 75.0 & 78.8 & \textbf{78.7} & 7.2 & 7.6 \\
        XLS-R E2E & 66.6 & 80.3 & 74.3 & 81.1 & 78.2 & 80.5 & 8.1 & 9.5 \\
        OWSM E2E & 53.6 & 78.5 & 60.7 & 92.1 & 81.0 & 117.1 & 14.0 & 11.3 \\
        \midrule
        \multicolumn{8}{l}{\textbf{Single task}} \\
        \cmidrule(r){1-1}
        WavLM E2E & 38.1 & \textbf{59.2} & 45.9 & \textbf{64.5} & 84.8 & 88.3 & 8.4 & 7.9 \\
        XLS-R E2E & \textbf{36.8} & 59.6 & \textbf{44.0} & 66.8 & 85.6 & 90.3 & 9.2 & 8.5 \\
        OWSM E2E & 48.2 & 67.7 & 54.8 & 80.0 & \textbf{75.0} & 102.9 & 13.7 & \textbf{11.6} \\
        \midrule
        \multicolumn{8}{l}{\textbf{Cascade}} \\
        \cmidrule(r){1-1}
        XLS-R + ByT5 & - & - & 48.5 & 70.6 & 86.7 & 124.1 & 16.0 & 11.0 \\
        XLS-R + ByT5 w/ ODIN & - & - & - & - & 85.5 & 120.8 & \textbf{16.6} & 10.6 \\
        \midrule
        \midrule
        \multicolumn{8}{l}{\textbf{Ground truth text}} \\
        \cmidrule(r){1-1}
        ByT5 & - & - & 16.0 & 28.1 & 55.2 & 157.0 & 22.0 & 12.2 \\
        ByT5 w/ ODIN & - & - & - & - & 47.7 & 137.2 & 23.0 & 12.2 \\
        \bottomrule
    \end{tabular}
    \caption{Results from multilingual experiments. The languages are split into two groups: ``seen'', where the languages are in the training set, and ``unseen'', where the languages are not in the training set. Each number represents an average of that metric across the languages in that group (macro-averaging). WavLM and XLS-R models have pretrained encoders while OWSM have pretrained encoder and decoder.}
    \label{tab:results}
\end{table*}

\section{Results} \label{sec:results}

We compare average model performance on seen languages (the 22 languages with training sets) and unseen languages (the 15 languages with only dev and test sets) in Table \ref{tab:results}. We report character error rates (CER, lower is better) for transcription, underlying, and glossing; and character F-scores (chrF++, \citealt{popovic-2017-chrf}, higher is better) for translation. We also evaluate translation outputs with BLEU \citep{papineni-etal-2002-bleu}, BLEURT \citep{sellam-etal-2020-bleurt}, and BERTScore (\citealt{Zhang2020BERTScore}, see Appendix \ref{apdx:translation}). All reported scores are macro-averaged across languages. We observe that multi-task models are worse across all tasks except glossing, where each of the two multi-task self-supervised models outperforms its single-task counterpart. Of the single-task end-to-end speech models, the two self-supervised models share similar performance across tasks, with the XLS-R based model performing best for transcription and underlying form prediction on seen languages. The OWSM model, on the other hand, is better at generating gloss and translation.

Since XLS-R is our best model for transcription, we employ it for the cascade setting so that its transcription outputs become the inputs for the text annotation model. The cascade approach produces better translation than all end-to-end models, but fails to improve on the underlying or glossing tasks.
Pretraining the text model on ODIN slightly improves glossing performance, but not enough to surpass end-to-end.

Unsurprisingly, models generally perform better on seen than unseen languages. 
The overall performance of the models is low in absolute numbers across most of the tasks, with ASR being the easiest task, and gloss and translation the hardest. 
Qualitatively, from inspecting the models' outputs on a few languages, it seems that they are not able to generate coherent and relevant translations. 
This highlights the challenges with building NLP resources for low-resource languages, with minimal data spread across many languages. 
Even models that were pretrained with multilingual datasets are hard to adapt to languages in \dataset{}.

\section{Discussion} \label{sec:discussion}
Some aspects of our experimental results highlight trends that may assist further development of \task{} technologies. 

\paragraph{Pre-trained vocabulary aids glossing and translation}
One notable difference between OWSM and the other two end-to-end speech models is that it includes a decoder pre-trained for transcription, translation, and language identification. 
Because the references for our gloss and translation tasks are in high-resource languages that were likely included in OWSM's large, multilingual training set, its BPE tokenizer has likely been exposed to many of their tokens. 
The references for our transcription and underlying prediction, however, are in low-resource languages likely not included in OWSM's training data. 
This phenomenon of tokenization is likely why OWSM performs comparatively well for glossing and trasciption, and comparatively poorly for transcription and underlying forms. 

\paragraph{Single-task beats multi-task}
We observe that single-task models are generally much better across all tasks, potentially because the tasks' diversity causes interference in multi-task objective settings \cite{yu2020gradient}. However, among pre-trained models, OWSM shows the smallest degradation from single- to multi-task performance, possibly because it was pre-trained with a multi-task objective and can thus extend better to new tasks.

\paragraph{Our cascaded models do not fully realize text-based potential}
From the results of the models trained on ground truth text in Table \ref{tab:results}, it is clear that annotation inference is easier from text than from speech. 
However, the advantage is not enough to overcome error propagation introduced by a cascade approach, at least for glossing. 
One appeal of text-input models is training data availability, since text IGT data is far more plentiful than speech. 
Our own results---indicating that fine-tuning on  noisy IGT (ODIN) improves glossing and translation---demonstrate some of this potential. For machine translation in particular, there are specialized pretrained models with much higher multilingual MT performance that could be deployed as part of the pipeline.
Novel approaches to this task should take this into account, and perhaps employ a multimodal model accepting both speech and text inputs.

\begin{figure}[h]
  \centering
  \includegraphics[width=0.9\columnwidth]{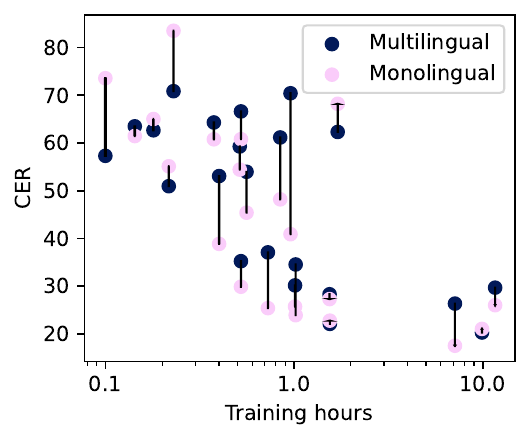}
  \caption{
  Comparing the seen language performance of multilingual transcription OWSM E2E model with monolingual models, trained separately with individual languages.
  Multilingual and monolingual performance is denoted as dark blue and light pink dots, respectively.
  Same language is connected with a black line.
  }
  \label{fig:mono_vs_multi}
\end{figure}

\paragraph{Multilinguality degrades performance except on the lowest-resource languages}
Multilingual training can vitally boost model performance for low-resource languages in some settings \citep{chen-etal-2019-multi-source}. 
However, multilinguality can also cause performance degradation for some languages, a phenomenon known as the ``curse of multilinguality'' \cite{conneau2020unsupervised}. 
Given this theoretical tension, we analyze monolingual versus multilingual training directly.  
We can only fairly compare OWSM models in this analysis, since self-supervised models rely on language models trained from scratch, and hence have different vocabulary sizes in monolingual vs. multilingual settings. 
Transcription error rates from end-to-end single-task OWSM models using monolingual versus multilingual training are shown in Figure \ref{fig:mono_vs_multi}. 
We observe that CER improves (decreases) as the dataset size increases. 
Generally, multilingual models perform better for small datasets, but this trend inverts as data increase, indicating multilingual performance degradation. 
This performance degradation is less pronounced for the highest-resource languages (Dolgan, Kamas, and Ainu; on the far right in Figure \ref{fig:mono_vs_multi}) than for other languages. 
We suspect that multilinguality is less harmful for these languages due to our checkpoint saving strategy. 
(See Section \ref{subsec:e2e}.) 
Because we keep the checkpoint with the best dev set accuracy in multilingual training, the checkpoints are biased towards languages with the highest dev set representation.

\section{Related Work}

Our study is informed by a richness of previous work in automatic glossing, including a number of proposed systems to predict IGT from segmented or unsegmented transcriptions. 
In the recent SIGMORPHON shared task on interlinear glossing, the best performing models included hard attention, transformer, and LSTM \citep{ginn-etal-2023-findings}. 
\citet{he-etal-2023-sigmorefun} showed that ByT5 models were not best at glossing, but had higher tolerance to noise and could benefit from noisy fine-tuning data such as ODIN (informing our decision to use them here).

\dataset{} is made possible by the work of field linguists. Previously, there have been many text-only IGT datasets from linguistic field works available for machine learning use, such as ODIN \citep{lewis-xia-2010}, IMTVault \citep{nordhoff-kramer-2022-imtvault}, or the SIGMORPHON IGT shared task data \citep{ginn-etal-2023-findings}.
However, to our knowledge, \dataset{} represents the first multilingual machine readable dataset focused on speech and interlinear gloss.

Our work is also informed by prior study in low-resource ASR. 
Previous low-resource ASR methodologies include fine-tuning high-resource ASR models \cite{cho2018multilingual} or self-supervised speech models \cite{baevski2020wav2vec,babu2021xls}. 
Later researchers built on these methods by
continuous pre-training \citep{tian22c_interspeech, chen20f_interspeech, metze2015semi, sakti2023leveraging},
model adaptation \citep{yu2023master, samarakoon2018domain},
and data augmentation \citep{khare21_interspeech, robinson2022tts}. 
These prior works informed our use of pre-trained models; however our work is the first to explore their adaptability to predict morphological, glossing, and translation annotations along with transcription. 
Since many low-resource languages need these annotations, \dataset{} can provide a valuable benchmark to facilitate future ASR and IGT prediction.

\section{Conclusion and Future Work}

In this work, we make a number of strides to advance technology-assisted documentation needed by language communities. 
We define a new task, \task{}, which is to predict IGT annotations directly from speech. 
We present \dataset{}, the first dataset for this task, comprised of audio files and expert annotations that were cleaned and formatted. 
We provide benchmarks across various training methodologies for transcription, underlying form prediction, glossing, and translation. 
And we analyze various prominent trends from experimental results. 
These data, benchmarks, and preliminary experimental insights provide a strong foundation for future \texttt{wav2gloss} breakthroughs, to expedite creation of needed language resources for communities of dying languages. 

This work may be continued in a number of ways. The models presented in this work represent baseline approaches, and we hope this work will spur on more experimentation on the optimal hyperparameters for this task. In particular, the choice of model size and the tokenization method can be further explored. Moreover, we think multi-modal setups such as \citet{barrault2023seamless} where both speech and text inputs can be fed into the model at the same time can be very promising for \task{}. Another possible avenue of research is the use of cascades consisting of more than two models, for example, ASR into morpheme segmentation into glossing.

Future researchers may also further normalize IGT labels, to expand \dataset{} to more diverse languages and phenomena. 
This is labor intensive, as previous research \citep{ListSimsForkel2021} and we ourselves observed, given inconsistent use of labels and language-specific phenomena across IGT collections. Expansion of \dataset{} to more languages may also come through community-driven projects to benefit low-resource language communities and academics. Researchers may also focus on improving our work's modeling capabilities---e.g., adapting models to zero-shot performance. In our own work, all models perform notably better on seen than on unseen languages. 
Future work may mitigate this by mapping all transcriptions to a shared vocabulary, such as IPA, to minimize superficial orthographic differences.

\section*{Limitations}

While we expect our contributions to be of significant value to the research community, we wish to acknowledge significant limitations to consider. 
Most notably to start, as is apparent from the scores in Table \ref{tab:results}, our models' performance in any of the four subtasks is not sufficient to render useful outputs for application. 
This highlights the challenging nature of working with low-resource languages and the novel \task{} task. 
By the release of our dataset and benchmarks, we mean to spur future iterations that improve upon our results and move towards applicable solutions for low-resource language communities. 

We acknowledge also that, since we did not perform linguistic annotations ourselves of the IGT datasets we include, our dataset's quality is tied to the accuracy of others' linguistic expertise and annotations. 
Since errors in annotation for training data, such as misalignments and mislabels, can cause mild to severe errors in machine learning outputs, we encourage users of our data to proceed with caution. 
We ensure our data's quality via cleaning and filtering, in addition to random manual inspection finding no severe errors, but we encourage continued vigilance in this vein. 

We hope this work can apply to other speech data from low-resource languages, since \dataset{} contains a diverse collection of languages, though we have not covered all possible writing systems and the effectiveness of the models with respect to rarer systems such as Cyrillic, Chinese, Arabic, etc.~is untested. We consider the systems we propose in this paper for research purposes only and have not tested generated texts for potential harmful or offensive contents.

People familiar with language resource archives will perhaps note the conspicuous absence of three of the largest archives of field linguistics -- TLA, ELAR, and PARADISEC. We forego these archives for this work mostly due to time and resource constraints, and will hopefully include them in the future. The challenges shared by the expansion to the three archives are mostly due to their sheer size and the work required to filter, extract, and obtain rights for the data. Most of this work will have to be done on a language-by-language basis to ensure quality.

\section*{Ethics Statement}

We wish to emphasize that any work in technologies for low-resource language communities should be approached with a high level of care for ethical practices. 
Many of these communities have particular needs and interests. 
And many have socioeconomic disadvantages that could be either helped or exacerbated by technological advancements. 

To be forthcoming about these important considerations, we wish to acknowledge some of the ethical concerns involved in the particular substance of our current work. 
We first acknowledge that data involved in this study were predominantly collected with the assumption that they would be used for language documentation and to support communities. 
While these are also our own end goals in developing the resources presented here, we recognize that researchers may use our open-source NLP technologies for a variety of purposes. 

We also wish to be straightforward about some potential ethical concerns with our data. 
While these data were collected with consent, in accordance with any pertinent legal and institutional protocols, they still contain sensitive materials. 
We did not manually anonymize the data, and therefore we urge caution to any users of \dataset{} to respect the rights and privacy of all individuals concerned as much as possible. 
We acknowledge that some low-resource language speakers may not see these technologies as a benefit and that they may be concerned about the potential effects of technology on their communities. 
We also express that we do not necessarily condone any opinions, worldviews, or assertions expressed within the transcriptions or recordings of our dataset. 
It is possible that some of their material may be offensive in some contexts. 
This is a common concern in building multilingual and multicultural datasets, since statements or references considered benign in one culture may be seen as offensive in another. 
We therefore caution users accordingly. 

Despite these potential concerns, which should be considered in all seriousness, we intend our work to have a highly positive effect, from an ethical standpoint. 
We hope our efforts can contribute to serving communities that have otherwise been left behind by many technological advancements, and to assist efforts to preserve valuable languages and cultures across the world, regardless of their socioeconomic privilege. 

\section*{Acknowledgements}
This work is supported by the US National Science Foundation grant \#2211951.
We also thank the linguists whose resources we used and their collaborators. Without them, this work would not have been possible.
For the computational resources, we used the Bridges2 system at PSC and Delta system at NCSA through allocation CIS210014 from the Advanced Cyberinfrastructure Coordination Ecosystem: Services \& Support (ACCESS) program, which is also supported by the US National Science Foundation grants \#2138259, \#2138286, \#2138307, \#2137603, and \#2138296.
Finally, we thank Paola Garcia, the anonymous reviewers, and the Area Chair for their valuable feedback to the present paper.

\bibliography{references/anthology,references/custom,references/language_resources}

\newpage
\appendix

\section{Parameter Counts and Hyper-parameters}
\label{apdx:params}
\begin{table}[h]
    \centering
    \begin{tabular}{lcc}
        \toprule
        \textbf{Model} & \textbf{Total} & \textbf{Trainable} \\
        \midrule
        WavLM E2E & 391M & 76M \\
        XLS-R E2E & 391M & 76M \\
        OWSM E2E & 101M & 101M \\
        ByT5 & 582M & 582M \\
        \bottomrule
    \end{tabular}
    \caption{Parameter counts of models used in this work.}
    \label{tab:parameters}
\end{table}

We show the parameter counts of models used in this work in Table \ref{tab:parameters}. All experiments are done with NVIDIA RTX A6000 GPUs. End-to-end models are trained with 4 GPUs and text models are trained with a single GPU. We used 1,605 GPU hours in total for both training and evaluation.

We do not perform extensive hyper-parameter searches. The settings we used across the experiments are shown in Table \ref{tab:hyperparams}.

\begin{table}[h]
    \centering
    \small
    \begin{tabular}{lccc}
        \toprule
         & \textbf{Conformer} & \textbf{OWSM} & \textbf{ByT5} \\
        \midrule
        Optimizer & Adam & AdamW & Adafactor \\
        LR & 2e-3 & 1e-3 & 5e-5 \\
        Warm up Steps & 25k & - & - \\
        Epochs & 30 & 10 & 10 \\
        \bottomrule
    \end{tabular}
    \caption{Hyper-parameter settings used in this study. ``Conformer'' includes both WavLM and XLS-R E2E models which share the same settings.}
    \label{tab:hyperparams}
\end{table}

\section{Additional Evaluation Metrics for Translation}
\label{apdx:translation}
\begin{table*}[t]
    \centering
    \begin{tabular}{l *{6}{c}}
        \toprule
        & \multicolumn{2}{c}{BLEU $\uparrow$} & \multicolumn{2}{c}{BLEURT $\uparrow$} & \multicolumn{2}{c}{BERTScore $\uparrow$} \\
        \cmidrule(lr){2-3} \cmidrule(lr){4-5} \cmidrule(lr){6-7}
        \textbf{Model} & Seen & Unseen & Seen & Unseen & Seen & Unseen \\
        \midrule
        \multicolumn{6}{l}{\textbf{Multi-task}} \\
        \cmidrule(r){1-1}
        WavLM E2E & 0.0 & 0.0 & -1.35 & -1.32 & 0.66 & 0.66 \\
        XLS-R E2E & 0.0 & 0.0 & -1.43 & -1.47 & 0.67 & 0.67 \\
        OWSM E2E & 2.1 & 0.1 & -1.26 & -1.36 & 0.72 & 0.69 \\
        \midrule
        \multicolumn{6}{l}{\textbf{Single task}} \\
        \cmidrule(r){1-1}
        WavLM E2E & 0.0 & 0.0 & -1.57 & -1.66 & 0.69 & 0.68 \\
        XLS-R E2E & 0.0 & 0.0 & -1.60 & -1.57 & 0.69 & 0.68 \\
        OWSM E2E & 2.0 & 0.1 & -1.29 & -1.43 & 0.72 & 0.69 \\
        \midrule
        \multicolumn{6}{l}{\textbf{Cascade}} \\
        \cmidrule(r){1-1}
        XLS-R + ByT5 & 2.7 & 0.0 & -1.27 & -1.52 & 0.73 & 0.69 \\
        XLS-R + ByT5 w/ ODIN & 2.9 & 0.0 & -1.27 & -1.52 & 0.74 & 0.69 \\
        \midrule
        \midrule
        \multicolumn{6}{l}{\textbf{Ground truth text}} \\
        \cmidrule(r){1-1}
        ByT5 & 6.6 & 0.2 & -1.00 & -1.48 & 0.77 & 0.70 \\
        ByT5 w/ ODIN & 6.7 & 0.3 & -1.00 & -1.48 & 0.77 & 0.70 \\
        \bottomrule
    \end{tabular}
    \caption{Additional metrics for translations.}
    \label{tab:transl-results}
\end{table*}
See Table \ref{tab:transl-results}. Results are obtained using the Evaluate\footnote{\url{https://github.com/huggingface/evaluate}} package.

\end{document}